\documentclass[conference]{IEEEtran}
\IEEEoverridecommandlockouts
\usepackage{amsmath,amssymb,amsfonts}
\usepackage{graphicx}
\usepackage{cite}
\usepackage{algorithmic}
\usepackage{textcomp}
\usepackage{threeparttable}
\usepackage{booktabs}
\usepackage[table]{xcolor}
\usepackage{makecell}
\usepackage{multirow}
\usepackage{soul}
\usepackage{url}
\usepackage{subfiles}
\usepackage{bbm}
\usepackage{tipa}
\usepackage{CJKutf8}

\begin{document}

\title{Cascade-Free Mandarin Visual Speech Recognition via Semantic-Guided Cross-Representation Alignment}

\author{\IEEEauthorblockN{Lei Yang, Yi He, Fei Wu, Shilin Wang\IEEEauthorrefmark{1}}
\IEEEauthorblockA{\IEEEauthorrefmark{1}Shanghai Jiao Tong University \\
\textit{\{yangleisx, heyi96, wu\_fei, wsl\}@sjtu.edu.cn}}
\thanks{Corresponding author.}}

\maketitle

\begin{abstract}

Chinese mandarin visual speech recognition (VSR) is a task that has advanced in recent years, yet still lags behind the performance on non-tonal languages such as English. One primary challenge arises from the tonal nature of Mandarin, which limits the effectiveness of conventional sequence-to-sequence modeling approaches.
To alleviate this issue, existing Chinese VSR systems commonly incorporate intermediate representations, most notably pinyin, within cascade architectures to enhance recognition accuracy. While beneficial, in these cascaded designs, the subsequent stage during inference depends on the output of the preceding stage, leading to error accumulation and increased inference latency.
To address these limitations, we propose a cascade-free architecture based on multitask learning that jointly integrates multiple intermediate representations, including phoneme and viseme, to better exploit contextual information. 
The proposed semantic-guided local contrastive loss temporally aligns the features, enabling on-demand activation during inference, thereby providing a trade-off between inference efficiency and performance while mitigating error accumulation caused by projection and re-embedding.
Experiments conducted on publicly available datasets demonstrate that our method achieves superior recognition performance.

\end{abstract}


\section{Introduction}
\label{sec:intro}

Visual speech recognition (VSR), which is also known as lipreading, is a technology that identifies the speech contents of a speaker solely using visual information from silent videos. It can be used to help people with hearing impairments in understanding videos, and also help speech recognition systems especially in high noise environments. The visual speech recognition methods have received extensive attention from both academic and industrial communities.
In recent years, with the rapid development of deep learning technologies, the VSR has made great progress. Especially, the data-driven end-to-end lipreading methods based on neural networks have been developed. Existing lipreading methods based on neural networks mainly include two stages: visual feature extraction and temporal feature aggregation. 

Early lipreading techniques primarily focus on word-level recognition. In the process of visual feature extraction, 2D Convolutional Neural Networks (CNN) and 3D CNN are widely used, as well as techniques such as Temporal Shift Module (TSM)~\cite{Hao2021tsm} from dynamic feature enhancement. In the temporal feature aggregation stage, techniques such as Temporal Convolution Network (TCN)\cite{Martinez2020tcn, Ma2021dctcn} and Recurrent Neural Network (RNN)\cite{Stafylakis2017combining, Petridis2017end} are adopted. Attention mechanisms are also widely adopted to extract temporal features~\cite{Chen2021automatic, Koumparoulis2022accurate}.
Sentence-level lipreading is more challenging. In sentence-level lipreading, contextual information is crucial for correctly recognizing the whole sentence.
The application of RNN~\cite{Chung2017lrs} and Transformer~\cite{Prajwal2022subword} and its variants, such as Conformer~\cite{Ma2021conformer, Fenghour2020onlyvisual, Ma2022visual, Burchi2023efficient} enhances the performance due to their contextual modeling capabilities. As for supervision, connectionist temporal classification (CTC) loss is commonly adopted, and hybrid CTC/Attention architecture~\cite{Watanabe2017hybrid} is being widely used.

\begin{figure}[t]
  \centering
  \includegraphics[width=\linewidth]{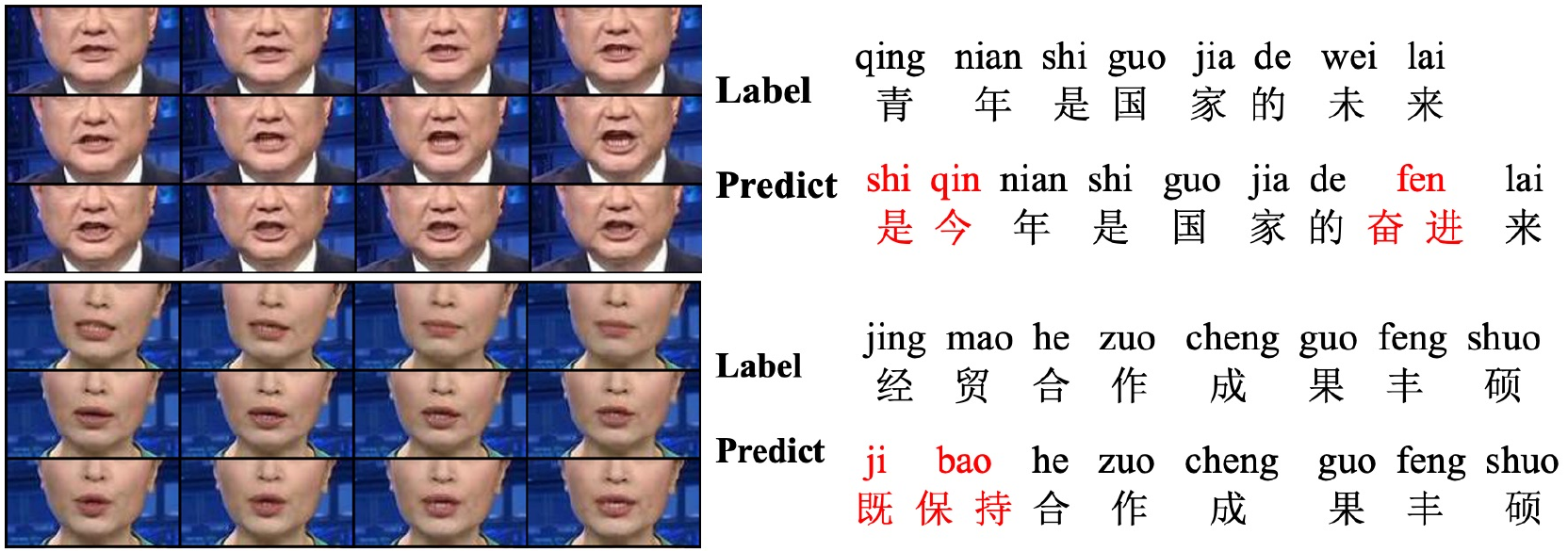}
  \caption{The error accumulation in cascade structures during inference. The error of predicted pinyin affects the prediction of characters, even with visual features in shortcut connection or cross level attention.}
  \label{fig:cascade}
\end{figure}

In Chinese mandarin VSR, which is a tonal language rather than phonetic language like English, there is a lack of direct correlation between Chinese characters and audio/visual signals. Moreover, the vocabulary of Chinese mandarin is large, which makes training the lipreading models more difficult.
A common solution is introducing intermediate representations that have better correlation with audio/visual signals in cascade structures. 
For example, Zhao et al.~\cite{Zhao2019CMLR} use pinyin syllables and tunes as intermediate representations in three cascaded sequence-to-sequence models based on GRU, 
Zhang et al.~\cite{Zhang2019lipch} use pinyin characters without blank in two-stage sequence-to-sequence models, which is Picture-to-Pinyin trained on visual modality and Pinyin-to-Hanzi trained on extra corpus,
Ma et al.~\cite{Ma2020ctch} introduce Transformer-based sequence-to-sequence model with pinyin characters as intermediate representation,
Ma et al.~\cite{Ma2022mir} use pinyin initials and finals, combined with viseme in cascaded structures, with the inference latency limited due to complicated cross stage attention.

However, the drawback of cascade structures is the error accumulation and inference latency due to strict stage-wise dependency. 
These methods require sequential execution of multiple stages during inference, 
the wrong result of the previous stage affects the next stage. As shown in Fig.~\ref{fig:cascade}, the error in pinyin prediction can affect the character prediction, where the character prediction results share a similar pronunciation with the incorrect pinyin. 
The projection and re-embedding operations between stages, as shown in Fig.~\ref{fig:compare}, also introduce additional errors and may lead to gradient discontinuities.
The strict stage-wise dependency also prevents parallel inference and increases latency, especially for real-time applications.

In this paper, a cascade-free architecture based on multitask learning is proposed to reduce error accumulation and improve the performance of Chinese mandarin VSR. We employ phoneme and viseme as intermediate representations due to their stronger associations with visual and acoustic features. During training, multiple intermediate representations are learned simultaneously to capture contextual information more effectively. During inference, intermediate representations are activated on-demand to achieve a trade-off between recognition performance and efficiency: visual features alone for faster inference, or combined with phoneme/viseme for better performance and interpretability. 
A semantic-guided local contrastive loss is constructed to temporally align phoneme and viseme representations. We adopt a phoneme-to-viseme mapping strategy as semantic guidance to ensure cross-representation consistency. Additionally, a local window alignment mechanism is employed to align intermediate representation sequences without relying on explicit temporal labels. The main contributions of this paper are as follows:

\begin{figure}
  \centering
  \includegraphics[width=0.9\linewidth]{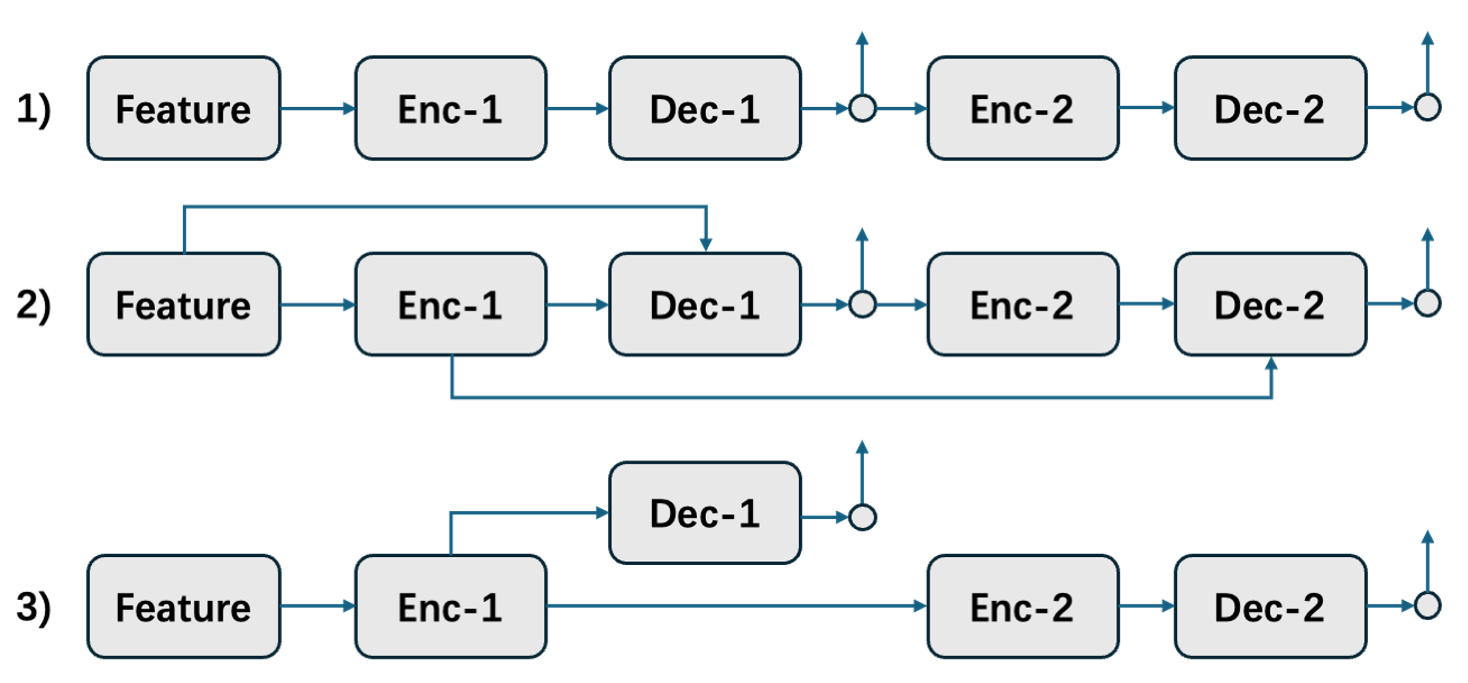}
  \caption{The comparison between multi-stage methods versus cascade-free method. 1) multi-stage method, 2) multi-stage method with shortcut connection, 3) cascade-free method.}
  \label{fig:compare}
\end{figure}

\begin{itemize}
  \item A novel cascade-free framework for Chinese mandarin VSR is proposed,
  which fundamentally eliminates the error accumulation caused by strict stage-wise dependency and projection/re-embedding operations in cascade structures. With multiple intermediate representations learned simultaneously and activated on-demand during inference, the framework can achieve a trade-off between recognition performance and inference efficiency.
  \item A semantic-guided local contrastive loss is constructed across different representations with semantic guidance to fully capture the contextual information. The cross-representation contrastive learning scheme improves the capabilities of the visual feature extractor for better temporal feature and aligns the intermediate representation sequences temporally for inference.
  \item The experimental results on public dataset CMLR demonstrate that the proposed method outperforms the state-of-the-art methods even without acoustic and additional information during training.
\end{itemize}

\section{The Proposed Method}
\label{sec:method}

\begin{figure*}[t]
  \centering
  \includegraphics[width=0.9\linewidth]{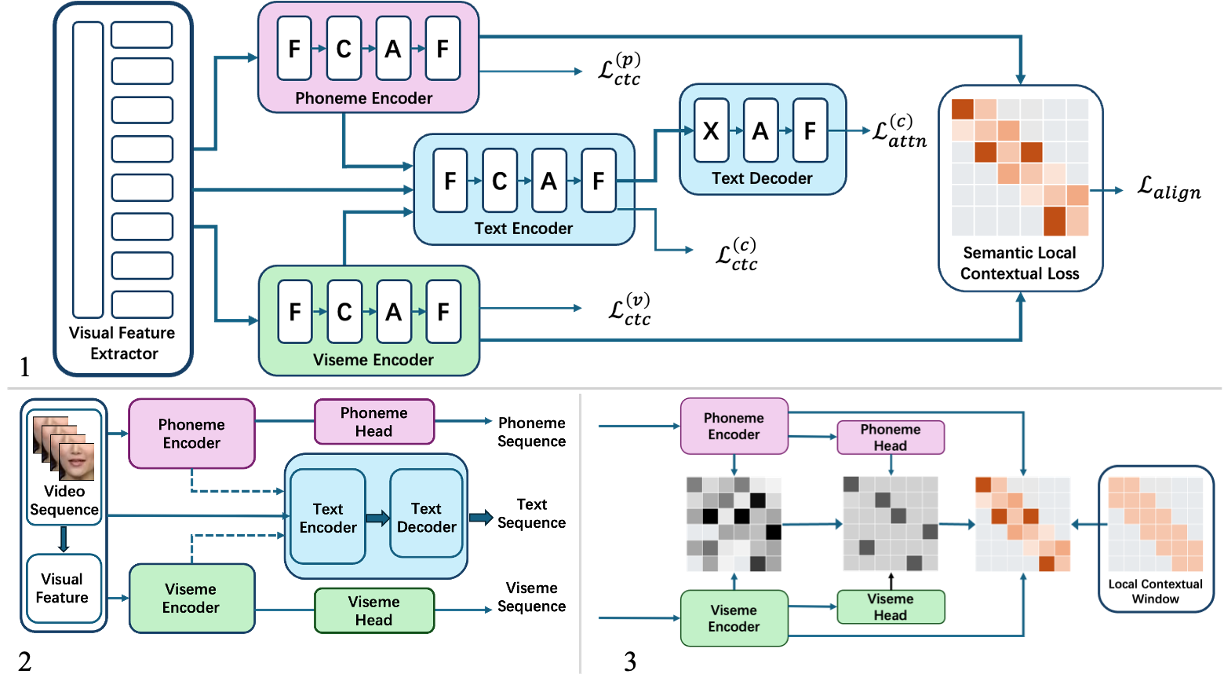}
  \caption{The overall architecture of the proposed method. 1) the training pipeline, F for feed-forward, C for convolution, A for self attention, X for cross attention. 2) the inference pipeline, the phoneme and viseme encoder could perform on-demand reasoning for recognition performance and interpretability. 3) the semantic-guided local contrastive loss, which aligns the intermediate representation temporally with semantic guide.}
  \label{fig:pipeline}
\end{figure*}

\subsection{Linguistic Settings}
\label{subsec:linguistic}


  
\begin{table}[!t]
  \centering
  \caption{The Viseme ID and its corresponding Phoneme symbols in IPA}
  \label{tab:phoneme}

  \begin{tabular}{lll}
    \toprule
    Viseme & Frequency &IPA\\
    \midrule
    0  & N/A     & \_  \\
    1  & 0.01\% & \textipa{P} \\
    2  & 3.08\%  & p, \textipa{p\super{h}}, m \\
    3  & 1.34\%  & f \\
    4  & 15.30\% & t, \textipa{t\super{h}}, n, l \\
    5  & 12.91\% & k, \textipa{k\super{h}}, x, \textipa{N} \\
    6  & 7.34\%  & \textipa{tC}, \textipa{tC\super{h}}, \textipa{C} \\
    7  & 8.00\%  & \textipa{\:t\:s}, \textipa{\:t\:s\super{h}}, \textipa{\:s}, \textipa{\:z} \\
    8  & 2.32\%  & ts, \textipa{ts\super{h}}, s \\
    9  & 8.81\%  & \textipa{A} \\
    10 & 11.43\% & e, o, \textipa{@}, \textipa{@\*r} \\
    11 & 15.56\% & \textipa{I}, \textipa{\:z}, \textipa{\textsyllabic{\:z}} \\
    12 &  7.81\% & \textipa{U} \\
    13 &  0.69\% & y \\
    14 &  2.36\% & \textipa{a\super{j}}, \textipa{e\super{j}} \\
    15 & 3.02\%  & \textipa{a\super{w}}, \textipa{o\super{w}} \\
  \bottomrule
\end{tabular}
\end{table}

In Chinese Mandarin, which is a tone-based non-phonetic language, different words may have the same or similar pronunciation with different tones. The correlation between characters and acoustic or visual signals is relatively limited.
However, in English and other phonetic languages, each letter or combination of letters represents a specific sound, and the order of letters in each word is highly related to pronunciation. Therefore, the correlation between characters and speech is relatively strong, which benefits the data-driven learning process of sequence-to-sequence models.
In this paper, phoneme and viseme are adopted as intermediate representations to bridge the gap between video frames and mandarin characters to assist the visual feature extraction.

Note that in this paper, the phoneme is employed rather than using pinyin character or initial-finals as previous works\cite{Zhao2019CMLR, Zhang2019lipch, Ma2020ctch, Ma2022mir}.
The phoneme constitute a transcription scheme based on the International Phonetic Alphabet (IPA), similar to the units used in Pinyin.
However, phonemes can provide representations that more closely reflect the actual process of human speech production\cite{odinye2015phonology, yang2024assessing}. 
In contrast, Pinyin, designed for ease of writing and widespread dissemination, has categorized and simplified many distinct pronunciations, making it difficult to represent subtle variations.

The viseme is the visual equivalent of phoneme, representing the visual articulation of speech sounds. Visemes group together phonemes that share similar lip movements and facial expressions during pronunciation.
The phoneme and viseme mapping scheme is shown in Table~\ref{tab:phoneme}. The linguistic-grounded mappings provide structured supervision for cross representation alignment, while the local contrastive loss described later still allow the model to learn the soft temporal correspondence within the local windows.
During preprocessing, pypinyin\footnote{ Public available in \url{https://github.com/mozillazg/python-pinyin}.} is employed to generate the viseme and phoneme sequence from the mandarin character labels.

\subsection{Multiple Intermediate Representation Learning}
\label{subsec:select}

The pipeline of the proposed method is shown in Fig.~\ref{fig:pipeline}.
The visual feature extractor is composed of two modules: local dynamic feature extraction and frame-wise feature extraction. Firstly, 3D conventional kernels are employed to capture local spatiotemporal dynamics. Secondly, a frame-wise feature extractor based on pretrained Res-Net is utilized to obtain visual representations. The visual feature $F \in \mathbf{R}^{B\times T \times C}$  is forwarded to the intermediate representation learning module.

Dedicated phoneme encoder and viseme encoder are employed to learn the representation from the visual features, each with an attached multilayer perceptron (MLP) head. The CTC loss is adopted as the supervisory signal to enforce intrinsic temporal alignment.
The intermediate representation learning strategy is designed to enhance the visual feature extractor by providing additional temporal context and alignment cues. Fine-grained linguistic labels from complementary perspectives enable the visual feature extractor to better capture the temporal dynamics of the input sequence, thereby improving performance on the downstream recognition task.

Given the visual features $F = [f_1, f_2, ..., f_t ]$ with the sequence length $t=1...T$, the phoneme representation $P = [p_1, p_2, ..., p_t ]$ and viseme representation $V = [v_1, v_2, ..., v_t ]$ are calculated through the phoneme encoder and viseme encoder.
To integrate multiple intermediate representation, a multi-branch DropPath fusion layer is employed to fuse the phoneme and viseme representation, which probabilistically masks and rescales the features before aggregation. The fusion is implemented as (\ref{eq:fusion}).
\begin{equation}
  \label{eq:fusion}
  \tilde{F} = \phi\left(F + (P \odot B_p) + (V \odot B_v)\right)
\end{equation}
where $B_p, B_v \sim \text{Bernoulli}(1- p_{drop})$ are independent stochastic masks for each branch and $\phi(\cdot)$ is the non-linear activation applied post-fusion.
This stochastic multi-branch regularization encourages robustness against feature redundancy, promotes complementary representation learning, and mitigates overfitting on any single modality.
\begin{align}\label{eq:pred}
    F_{mem} &= \text{Encoder}^{(c)}(\tilde F) \\
    \hat{y}_{attn} &= \text{MLP}_{attn} \left( \text{Decoder}^{(c)}(F_{mem}) \right) \\
    \hat{y}_{ctc} &= \text{MLP}_{ctc} ( F_{mem} )
\end{align}
The loss function is the hybrid CTC/Attention loss with $\alpha$ as weighting factor.
\begin{align}\label{eq:text-loss}
    \mathcal{L}_{ctc}^{(c)} &= -\log p(L_c | \hat{y}_{ctc}) \\
    \mathcal{L}_{attn}^{(c)} &= \text{CrossEntropyLoss} (L_c, \hat{y}_{attn}) \\
    \mathcal{L}^{(c)} &= \alpha \cdot \mathcal{L}_{attn}^{(c)} + (1-\alpha) \cdot \mathcal{L}_{ctc}^{(c)}
\end{align}

\subsection{Semantic-Guided Local Contrastive Loss}
\label{subsec:context}

To enable the visual feature extractor to capture features that are highly relevant to the intermediate representation, we construct a semantic local contrastive loss that focuses on local context. The semantic local contrastive loss combines contrastive learning and semantic mapping constraints to capture semantic consistency while maintaining locality in the temporal dimension.

Given the extracted viseme features $ V = [v_1, v_2, ..., v_T ]$ and phoneme features $ P = [p_1, p_2, ..., p_T ]$ with the sequence length (number of frames) $T$, firstly, the mapping matrix $M$ is constructed based on the semantic mapping relationships between predicted logits of phoneme $\hat y^{(p)}_{j}$ and viseme $\hat y^{(v)}_{i}$ as shown below.
\begin{equation}
\label{eq:map}
    M_{v_i, p_j} =
    \begin{cases}
        1, & \text{If } \hat y^{(p)}_{j} \sim \hat y^{(v)}_{i}  \\
        0, & \text{otherwise}
    \end{cases}
\end{equation}
In order to promote the corresponding learning of temporally adjacent tokens, local window mask as (\ref{eq:window}) is introduced.
\begin{equation}
    \label{eq:window}
    W_{i,j} =
    \begin{cases}
    1, & |i - j| \leq r \\[4pt]
    0, & \text{otherwise}
    \end{cases}
\end{equation}
Where $r = \lfloor \frac{w}{2} \rfloor$, is the half window width. This mask ensures that each viseme token \(\mathbf{v}_i\) only contrasts with phoneme tokens within the window \([i-r, i+r]\).
The features are L2 normalized and the cosine similarity matrix is calculated as (\ref{eq:sim}).
\begin{equation}
    \label{eq:sim}
    S_{i,j} = \frac{\mathbf{v}_i^T \mathbf{p}_j}
    {|| \mathbf{v}_i || || \mathbf{p}_j ||}
\end{equation}
Semantics $M_{v_i, p_j}$ and local windows$W_{ij}$ jointly determine the positive sample mask as (\ref{eq:pos}).
\begin{equation}\label{eq:pos}
P_{i,j} = \mathbbm{1} \left (M_{y^{(v)}_i, y^{(p)}_j} = 1\right ) \cdot W_{i,j}
\end{equation}
The similarity is temperature scaled \( \tau \) to form a local distribution as (\ref{eq:dist}).
\begin{equation}\label{eq:dist}
q(j|i) = \frac{\exp(S_{i,j} / \tau)}{\sum_{k: W_{i,k}=1} \exp(S_{i,k}/\tau)}
\end{equation}
The positive sample distribution is defined as (\ref{eq:pos-dist}).
\begin{equation}\label{eq:pos-dist}
p(j|i) = \frac{P_{i,j}}{\sum_{k} P_{i,k}}
\end{equation}
The final semantic local contrast loss, as (\ref{eq:loss}), is the KL divergence expectation for each sample.
\begin{equation}
\label{eq:loss}
\mathcal{L}_{align} = \frac{1}{B} \sum_{b=1}^{B}
\frac{
    \sum_{i} \text{KL} \left ( p_b(\cdot|i) \parallel q_b(\cdot|i) \right ) 
    \cdot \mathbbm{1}[\sum_{j} P_{b,i,j} > 0]}
{\sum_{i} \mathbbm{1}[\sum_{j} P_{b,i,j} > 0] + \epsilon}
\end{equation}
The total training loss is as follows, where $\lambda_1$ and $\lambda_2$ are hyperparameters to balance the losses.
\begin{equation}
\mathcal{L} = \mathcal{L}^{(c)} + \lambda_1 \cdot \mathcal{L}_{align} + \lambda_2 \cdot \left(\mathcal{L}_{ctc}^{(p)} + \mathcal{L}_{ctc}^{(v)}\right)
\end{equation}

\subsection{On-demand Activation Inference}
\label{subsec:inference}

During inference, the on-demand activation provides a controllable trade-off between accuracy and efficiency, which allows selecting which intermediate representations to use: visual features alone for faster inference, or combined with phoneme/viseme for higher accuracy.
The cascade-free architecture eliminates the strict stage-wise dependency, allowing for parallel processing of different branches. This design significantly reduces inference latency compared to traditional cascaded structures, where each stage must wait for the previous one to complete.

\section{Experiments}
\label{sec:expr}

\subsection{Dataset and Evaluation Metrics}
\label{subsec:expr-dataset}

\begin{table}[!t]
  \caption{Dataset Details. U for utterances and T for time}
  \label{tab:dataset}
  \begin{tabular}{l|cc|cc}
    \toprule
    \multirow{2}{*}{Dataset} & \multicolumn{2}{c|}{Seen} & \multicolumn{2}{c}{Unseen}\\
    & \# U / T & Max Frames & \# U / T & Max Frames \\
    \midrule
    Train & 71448 / 60.6h & 223 & 64879 / 54.9h & 208 \\
    Valid & 10206 / 8.6h & 208 & 16224 / 13.8h & 216 \\
    Test & 20418 / 17.3h & 223 & 20980 / 17.8h & 223 \\
  \bottomrule
\end{tabular}
\end{table}

To evaluate the performance of proposed network, experiments are conducted on the publicly available CMLR dataset~\cite{Zhao2019CMLR}.
CMLR is a public Chinese Mandarin lipreading dataset, consisting of over 100,000 sentences spoken by 11 speakers. 
Experiments were conducted in both seen and unseen scenarios. In the seen scenario, the official 7:1:2 data split was used. In the unseen scenario, the utterances from s2 and s6 are used for testing and the others are adopted for training. 
The details of the dataset are demonstrated in Table~\ref{tab:dataset}.
As an evaluation metric, the Character Error Rate, as $CER = \frac{S+D+I}{N}$, is calculated on the predicted sequence (hypothesis) and the ground truth (reference).
The $S$ is the number of substitutions, the $D$ is the number of deletions, the $I$ is the number of insertions from the reference to the hypothesis and the $N$ is the length of the reference sequence.

\subsection{Implementation Details}
\label{subsec:expr-setting}

In the experiments conducted on the CMLR dataset, the character encoder consists of 6 layers, while the character decoder contains 3 layers.
The phoneme and viseme encoders are both 3 layers. The local window width is set to 5 according to the average duration of each viseme.
For all experiments, the AdamW optimizer and Cosine Annealing scheduler with warmup is adopted.
The curriculum learning strategy is adopted, all the models are firstly trained for 50 epochs on the subset that is less than 64 frames, with a maximum learning rate 1e-3. Then the models are trained with the whole training set, with a maximum learning rate 1e-4.
For data augmentation, random cropping is adopted in both stages, and random time masking is adopted in the second stage.

\subsection{Performance Comparison}
\label{subsec:expr-comparison}

\begin{table}[!t]
    \centering
    \caption{Performance Comparison on CMLR dataset. Modality: V for video, A for audio, L for facial landmarks.}
    \label{tab:sota-cmlr}

    \begin{tabular}{lccc}
      \toprule
      Method & Modality & CER\%,~Seen~$\downarrow$ & CER\%,~Unseen~$\downarrow$ \\
      \midrule
      Lipnet~\cite{Assael2016lipnet}        & V   & 33.41 & 52.18 \\
      WAS~\cite{Chung2017lrs}               & V   & 38.93 & -     \\
      LipCH-Net~\cite{Zhang2019lipch}       & V   & 34.07 & -     \\
      CSSMCM~\cite{Zhao2019CMLR}            & V   & 32.48 & 50.08 \\
      CTCH-LipNet~\cite{Ma2020ctch}         & V   & 22.02 & 62.47 \\
      CT-MIR-Net~\cite{Ma2022mir}           & V   & 21.45 & 51.38 \\
      LCSNet~\cite{Xue2023lcsnet}           & V   & 30.03 & 46.98 \\
      \midrule
      LIBS~\cite{Zhao2020libs}              & A+V & 31.27 & -     \\
      CALLip~\cite{Huang2021callip}         & A+V & 31.18 & -     \\
      \midrule
      LipFormer~\cite{xue2023lipformer}     & V+L & 27.79 & 43.18 \\
      GusLip ~\cite{Li2024guslip}           & V+L & 29.98 & 43.61 \\
      \midrule
      Proposed  & V & \textbf{20.38} & \textbf{38.23} \\
      \bottomrule
    \end{tabular}
\end{table}

The experimental results on CMLR dataset are presented in Table~\ref{tab:sota-cmlr}.
The proposed method achieves better performance across both seen and unseen evaluation scenarios. Under the video-only modality, the proposed model attains error rates of 20.38\% under the seen settings and 38.23\% under the unseen settings, outperforming all competing approaches. In particular, it surpasses prior video-based architectures such as Lipnet~\cite{Assael2016lipnet}, WAS~\cite{Chung2017lrs}, and LCSNet~\cite{Xue2023lcsnet}, showing notable reductions in error.

Furthermore, despite relying solely on video input, the proposed method also outperforms several systems with auxiliary input modalities, such as LipFormer~\cite{xue2023lipformer} and GusLip~\cite{Li2024guslip}, both of which incorporate additional facial landmark information. These results indicate that the proposed architecture effectively captures discriminative visual cues without requiring supplementary modalities.
Overall, the superior performance across all settings verifies the robustness and generalization capability of the proposed network.

\subsection{Ablation Study}
\label{subsec:expr-ablation}

\begin{table}[!t]
  \centering
  \caption{Ablation Study on CMLR dataset}
  \label{tab:ablation}

  \begin{tabular}{lcc}
    \toprule
    model & CER\%,~Seen~$\downarrow$ & CER\%,~Unseen~$\downarrow$ \\
    \midrule
    Proposed & 20.38 & 38.23 \\
    \quad w/o $\mathcal{L}_{align}$  & 24.34 & 42.13 \\
    \quad \quad w/o phoneme and viseme & 35.25 & 49.78 \\
    \bottomrule
  \end{tabular}
\end{table}

The ablation study results in Table~\ref{tab:ablation} demonstrate the effectiveness of each component. The semantic-guided local contrastive loss is crucial for the proposed network, which effectively enhances the model's discriminative ability by temporally aligning intermediate representations with semantic guidance. With this loss, the visual feature extractor captures more speech production-related information, improving CER from 24.34\% to 20.38\% (Seen) and from 42.13\% to 38.23\% (Unseen).

Removing both phoneme and viseme supervision causes substantial degradation: CER rises to 35.25\% (Seen) and 49.78\% (Unseen). This single-stage sequence-to-sequence model (video-to-characters) performs significantly worse, confirming that intermediate phoneme/viseme representation learning is essential for robust visual speech recognition.

\subsection{Intermediate Representation Effectiveness}
\label{subsec:expr-select}

\begin{table}[!t]
  \centering
  \caption{Activation Parameters and Recognition Performance}
  \label{tab:select}

  \begin{tabular}{lccc}
    \toprule
    features & CER\%,~Seen~$\downarrow$ & CER\%,~Unseen~$\downarrow$ & \#Active Params\\
    \midrule
    $F$     & 24.47 & 45.72 & 86.79M \\
    $F+P$   & 22.61 & 40.15 & 108.20M \\
    $F+V$   & 24.37 & 41.37 & 108.18M\\
    $F+P+V$ & 20.38 & 38.23 & 129.58M \\
    \bottomrule
  \end{tabular}
\end{table}

We evaluate the effectiveness of different intermediate representations. Results in Table~\ref{tab:select} show that multitask learning enhances visual feature extraction. The model with only visual features achieves 24.47\% (Seen) and 45.72\% (Unseen) CER, comparable to models without intermediate representation. Incorporating both phoneme and viseme features achieves the best performance: 20.38\% (Seen) and 38.23\% (Unseen) CER. Viseme provides limited improvement in Seen scenarios but better robustness in Unseen scenarios, indicating speaker-independent features.

\begin{table}[!t]
  \centering
  \caption{Inference Efficiency Comparison}
  \label{tab:inference}

  \begin{tabular}{lccc}
    \toprule
    Method & Structure & \#Params & Infer(ms) \\
    \midrule
    Lipnet~\cite{Assael2016lipnet} & single-stage, LSTM & 18.16 M & 52.3  \\
    CSSMCM~\cite{Zhao2019CMLR}     & three-stage, LSTM & 38.45 M & 137.2 \\
    \midrule
    CTCH-Lipnet~\cite{Ma2020ctch}  & dual-stage, Transformer & 40.14 M & 92.7 \\
    CT-MIR-Net~\cite{Ma2022mir}    & dual-stage, Transformer & 122.57 M & 160.9 \\
    \midrule
    $F$     & cascade-free, Conformer & 86.79 M  & 79.1 \\
    $F+P$   & cascade-free, Conformer & 108.20 M & 87.6 \\
    $F+V$   & cascade-free, Conformer & 108.18 M & 84.7 \\
    $F+P+V$ & cascade-free, Conformer & 129.58 M & 95.2 \\
    \bottomrule

  \end{tabular}
\end{table}

Table~\ref{tab:inference} compares inference efficiency. LSTM-based models show linear latency growth with cascade stages due to sequential processing. Multi-stage Transformer approaches suffer from inter-stage dependencies, where beam search in earlier stages must complete before subsequent stages proceed. In contrast, the proposed cascade-free Conformer eliminates inter-stage dependencies, achieving lower inference latency while maintaining competitive capacity.

\begin{table}[!t]
  \centering
  \caption{Inference Quality with Different Activated Intermediate Representation}
  \label{tab:quality}

  \begin{tabular}{l|l|c}
    \toprule
    Ground Truth &
    \begin{CJK}{UTF8}{gbsn}
      国务院督察组将督促整改
    \end{CJK}
    & CER \\
    \midrule
    F &
    \begin{CJK}{UTF8}{gbsn}
      国务院督\textcolor{red}{查}组将\textcolor{red}{陆续展开}
    \end{CJK}
    & 0.4545 \\
    F+P &
    \begin{CJK}{UTF8}{gbsn}
      国务院督\textcolor{red}{查}组将\textcolor{red}{突出}整改
    \end{CJK}
    & 0.2727 \\
    F+V &
    \begin{CJK}{UTF8}{gbsn}
      国务院督\textcolor{red}{查}组将\textcolor{red}{图书}整改
    \end{CJK}
    & 0.2727 \\
    F+P+V &
    \begin{CJK}{UTF8}{gbsn}
      国务院督\textcolor{red}{查}组将督促整改
    \end{CJK}
    & 0.0909 \\
    \bottomrule
  \end{tabular}

\end{table}

\subsection{Discussion}
\label{subsec:discuss}

Some homophones still exist in the prediction results.
As shown in Table~\ref{tab:quality}, even with all intermediate representations activated (F+P+V), some homophone errors persist (e.g., misrecognizing similar syllables). This suggests that while phoneme and viseme features improve recognition, certain semantically ambiguous characters require additional contextual reasoning beyond local acoustic-visual correspondence.
Mandarin is a non-phonetic language where phonetically similar syllables with different tones convey distinct meanings, posing challenges for visual-only recognition. Contextual information is a promising direction for homophone disambiguation.
The performance gap between unseen and seen speakers is a common issue in VSR. Future work could explore identity disentanglement techniques to improve generalization to unseen speakers.

\section{Conclusion}
\label{sec:conclusion}

In this paper, we presented a cascade-free framework for Mandarin VSR that avoids the limitations inherent in cascade-based designs.
By jointly incorporating phoneme and viseme as complementary intermediate representations, the proposed approach leverages their stronger correlation with visual-acoustic patterns while eliminating repeated projection and re-embedding operations that often introduce error accumulation.
The semantic-guided local contrastive learning mechanism further enhances temporal alignment across representations, enabling the model to capture richer contextual cues essential for accurate sentence-level prediction.
Experimental results verify that our method delivers better performance than existing approaches, particularly in unseen scenarios, and offers a promising direction for future research in Mandarin VSR.

\bibliographystyle{IEEEtran}
\bibliography{refs}

\end{document}